\documentclass[letterpaper, 10 pt, conference]{ieeeconf}  % Comment this line out if you need a4paper

\IEEEoverridecommandlockouts                              % This command is only needed if 
\usepackage[utf8]{inputenc} % allow utf-8 input
\usepackage[T1]{fontenc}    % use 8-bit T1 fonts
\usepackage{microtype}

\usepackage{graphicx}
\usepackage{subcaption}
\usepackage{amsmath,amssymb,mathtools}
\usepackage{bm}
\usepackage{algorithm}
\usepackage[noend]{algpseudocode}
\usepackage[colorlinks=true,citecolor=blue,linkcolor=blue]{hyperref}
\usepackage[nameinlink,capitalize,noabbrev]{cleveref}
\usepackage{xspace}
\usepackage{cite}
\usepackage{tikz}

\usepackage{booktabs,tabularx,makecell}
\newcolumntype{Y}{>{\centering\arraybackslash}X}

\usepackage{multirow}
\usepackage{siunitx}

\usepackage{rotating}
\let\labelindent\relax
\usepackage{enumitem}
\usetikzlibrary{arrows.meta,positioning,fit,calc,shapes,backgrounds}
\usepackage{xcolor}

\newcommand{\bs}{\boldsymbol}

\newcommand{\vhd}{\textbf{Variable Horizon Diffuser}\xspace}
\newcommand{\method}{\textbf{VHD}\xspace}
\newcommand{\lenpred}{\textbf{Length Predictor}\xspace}
\newcommand{\diffplan}{\textbf{Diffusion Planner}\xspace}

\newcommand{\mrow}[1]{\multirow{2}{*}[-0.6ex]{\strut #1}}

\title{\LARGE \bf
VH-Diffuser: Variable Horizon Diffusion Planner for\\ Time-Aware Goal-Conditioned Trajectory Planning
}

\author{Ruijia Liu, Ancheng Hou, Shaoyuan Li and Xiang Yin% <-this % stops a space
\thanks{R. Liu, A. Hou, S. Li and  X. Yin are with the Department of Automation, Shanghai Jiao Tong University, Shanghai 200240, China. {\tt  E-mail: \{liuruijia,hou.ancheng,syli,yinxiang\}@sjtu.edu.cn}} 
}

\begin{document}

\maketitle
\thispagestyle{empty}
\pagestyle{empty}
\setlength{\abovecaptionskip}{0pt}
\setlength{\belowcaptionskip}{0pt}
\setlength{\textfloatsep}{6pt}

%%%%%%%%%%%%%%%%%%%%%%%%%%%%%%%%%%%%%%%%%%%%%%%%%%%%%%%%%%%%%%%%%%%%%%%%%%%%%%%%
\begin{abstract}
Diffusion-based planners have gained significant recent attention for their robustness and performance in long-horizon tasks. However, most existing planners rely on a fixed, pre-specified horizon during both training and inference.
This rigidity often produces length–mismatch (trajectories that are too short or too long) and brittle performance across instances with varying geometric or dynamical difficulty. 
In this paper, we introduce the \vhd (\method) framework, which treats the horizon as a learned variable rather than a fixed hyperparameter. Given a start-goal pair, we first predict an instance-specific horizon using a learned \lenpred model, which guides a \diffplan to generate a trajectory of the desired length. Our design maintains compatibility with existing diffusion planners by controlling trajectory length through initial noise shaping and training on randomly cropped sub-trajectories, without requiring architectural changes. Empirically, \method improves success rates and path efficiency in maze-navigation and robot-arm control benchmarks, showing greater robustness to horizon mismatch and unseen lengths, while keeping training simple and offline-only.
\end{abstract}

% =========================================================
% \section{Introduction}
% \label{sec:intro}
% Diffusion-based planners have recently emerged as a robust and versatile tool for offline planning and control. However, most existing approaches operate at a \emph{fixed horizon}, typically chosen via validation or by training a grid of models with different horizons. This rigidity introduces three key limitations: (i) \emph{length mismatch}—plans that are too short fail to reach the goal, while overly long plans waste steps or diverge; (ii) \emph{poor scalability}—covering a wide range of horizons requires either multiple models or coarse post hoc selection; and (iii) \emph{dataset skew}—offline trajectories often contain dithering or detours, making trajectory length a noisy and unreliable supervision signal for generative models.

\section{Introduction}
\label{sec:intro}
Diffusion-based planners have recently emerged as robust and versatile tools for offline planning and control \cite{janner2022planning,ajay2022conditional,chi2023diffusion,carvalho2023motion,xiao2025safe,huang2025diffusion,feng2025diffusion,zhou2025emdm,liu2025zero}. The idea of diffusion-based planning is to learn a generative model of trajectories from offline data and sample trajectories that satisfy desired task constraints while adhering to dynamic constraints via conditional denoising during inference. This approach offers significant advantages, such as the ability to handle long-horizon tasks effectively and produce high-quality, smooth plans in complex, high-dimensional spaces.

\begin{figure}[t]
  \centering
  \includegraphics[width=0.48\textwidth]{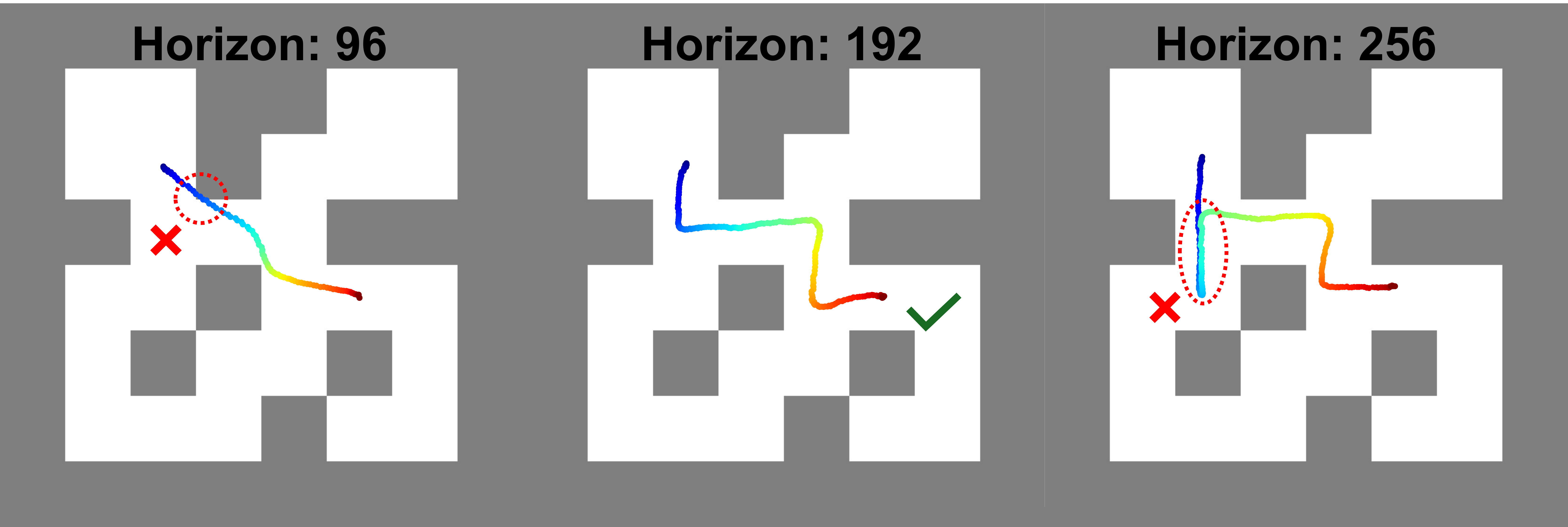}
  \caption{\textbf{Length mismatch in fixed-horizon diffusion planning.}
  For the same start and goal, we plot trajectories planned with horizons $H\!\in\!\{96,192,256\}$. \textbf{Left} ($H{=}96$): the horizon is too short and the plan is infeasible. \textbf{Middle} ($H{=}192$): the horizon matches the instance, yielding a concise feasible plan. \textbf{Right} ($H{=}256$): the horizon is too long, inducing detours.}
  \label{fig:mismatch}
\end{figure}

However, most existing diffusion-based planners rely on a \emph{fixed horizon}, typically chosen based on prior experience or validation. This fixed horizon can be brittle at the instance level. Fig.~\ref{fig:mismatch} illustrates the issue for a single start–goal pair: a horizon of $H{=}96$ under-shoots the goal (left), $H{=}192$ aligns with the instance, producing a concise, near-geodesic plan (middle), while $H{=}256$ over-shoots the goal and introduces detours and dithering (right). We refer to this misalignment between a preset horizon and the geometric or dynamical difficulty of a specific case as \emph{length mismatch}.

Length mismatch has broader consequences: (i) it compromises reliability, as too-short plans may fail to reach the goal, while overly long plans waste steps or even diverge; (ii) it limits scalability, as covering a wide range of horizons requires either multiple models or inefficient post hoc selection; and (iii) it interacts poorly with offline datasets, where trajectories often exhibit dithering and detours, making length a noisy and unreliable supervision signal for generative models. These challenges motivate treating the horizon not as a fixed hyperparameter, but as an \emph{instance-specific} quantity that can be learned and adaptively controlled at test time.

In this work, we address these issues by \emph{decoupling} ``when to stop" from ``how to move". 
Specifically, our proposed \vhd (\method) framework consists of two cooperative modules, both trained on the same offline dataset: 
(1) a \lenpred module that estimates the shortest-step distance between states, and 
(2) a \diffplan module that generates a trajectory of the specified length. 
Unlike length-conditioned encoders, our approach preserves the planner architecture and instead \emph{controls trajectory length by modifying the initial noise trajectory length}. This design offers a minimal-code pathway to variable-length planning while ensuring   compatibility with existing diffusion backbones.

Our main contributions are as follows:
\begin{itemize}[leftmargin=*]
  \item We introduce \method, a modular framework that predicts per-instance horizons and enables variable-length diffusion planning with negligible architectural modification.
  \item We develop a practical length predictor model that estimates shortest-step distances between start-goal pairs. The model is trained using anchored supervision, Bellman-style upper bounds, and relay-based triangle regularization within a normalized target space.
  \item  
  We propose a simple yet effective \emph{random-length training} strategy for diffusion planning: training with randomly cropped sub-trajectories, while inference controls horizon through the length of the initial noise. This mechanism substantially improves length generalization.
  \item We present comprehensive experiments comparing against fixed-horizon diffusion planners, along with analyses of robustness to horizon mismatch.
\end{itemize}

% =========================================================
\section{Related Work}
\label{sec:related}
\textbf{Diffusion models for planning and variable-length generation.}
Diffusion model\cite{ho2020denoising} has become a strong substrate for offline planning by learning trajectory distributions and sampling plans under task constraints, representative works including \emph{Diffuser} for value-guided planning, \emph{Decision Diffuser} for conditional decision making, and \emph{Diffusion Policy} for visuomotor control~\cite{janner2022planning,ajay2022conditional,chi2023diffusion}. While these frameworks typically operate with a \emph{fixed} horizon (or explicit time tokens), \cite{janner2022planning} notes that one can generate variable-length plans by changing the initial noise length; however, this capability has not been systematically studied. Our framework develops this direction by pairing a learned \emph{Length Predictor} that estimates an instance-specific shortest-step horizon with a standard \emph{Diffuser} backbone \emph{without} architectural changes, realizing length control solely via the noise shape, and training the planner on \emph{random-length} trajectory crops to endow genuine length generalization.

\textbf{Learning distances and geodesics.}
There has been extensive prior work on learning temporal distances, goal-conditioned value functions, and related notions of distance-to-goal~\cite{venkattaramanujam2019self,hartikainen2019dynamical,ma2022vip,vasilev2024dynamical,myers2024learning}. Our Length Predictor differs in both objective and supervision: we target \emph{shortest-step} distance as a geometric reachability surrogate and learn from offline trajectories using hybrid supervisions—exact \emph{anchors} for intra-trajectory pairs, dynamic-programming upper bounds for $k$-step links, and triangle-type relay constraints—without requiring reward signals or explicit actions. This design biases estimates toward conservative (shortest) horizons and supports cross-trajectory stitching from purely observational data.

% =========================================================
\section{Problem Setup}
\label{sec:setup}

We consider an unknown discrete-time dynamical system:
\begin{equation}
s_{t+1} \;=\; f(s_t, a_t), \quad t=0,1,2,\dots,
\end{equation}
with state space $\mathcal{S}\subseteq\mathbb{R}^n$, action space $\mathcal{A}\subseteq\mathbb{R}^m$, and an unknown transition function $f:\mathcal{S}\times\mathcal{A}\to\mathcal{S}$. A (finite) \emph{trajectory} of length $T\in\mathbb{N}_{\ge 0}$ is a sequence $\bs{\tau}=(s_0,a_0,s_1,a_1,\dots,a_{T-2},s_{T-1})$ containing $T-1$ steps of transitions
such that $s_{t+1}=f(s_t,a_t),\forall t=0,\dots,T-2$. 

We are given an offline trajectory dataset of $N$ episodes
\[
\mathcal{D} \,=\, \bigl\{\,(s_t^i,a_t^i,s_{t+1}^i)\,\bigr\}_{t=0}^{T_i-1},\qquad i=1,\dots,N,
\]
collected under an unknown behavior policy and not assumed to be optimal. Thus, the episodes may include dithering, detours, and other suboptimal artifacts.

\textbf{Start-Goal Specifications and Goal Regions.}
For a start-goal pair $(s,g)\in\mathcal{S}\times\mathcal{S}$, we fix a tolerance threshold $\varepsilon>0$ and define the goal set as
\[
\mathcal{G}_\varepsilon(g)\,=\,\bigl\{\,s'\in\mathcal{S}\; \big|\; \lVert s'-g\rVert_\infty \le \varepsilon\,\bigr\}.
\]
A trajectory $\tau$ \emph{reaches} $g$ in $k$ steps if $s_0=s$ and $s_k\in \mathcal{G}_\varepsilon(g)$.

\textbf{Shortest-Step Distance.}
The \emph{shortest-step distance} (optimal horizon) between $s$ and $g$ is
\begin{equation}
D^\star(s,g)
= \inf\left\{\,k\in\mathbb{N}_{\ge 0}\;\middle|\;
\begin{aligned}
& \exists\,\tau \text{ of length } k+1, \\
& s_0 = s,\;\; s_k \in \mathcal{G}_\varepsilon(g)
\end{aligned}
\right\}.
\label{eq:shortest-horizon}
\end{equation}
If $g$ is unreachable from $s$ then $D^\star(s,g)=+\infty$. In practice we cap horizons by $T_{\max}\in\mathbb{N}$ and use the truncated distance
\begin{equation}
D_{T_{\max}}(s,g)\,=\,\min\{\,D^\star(s,g),\; T_{\max}\,\}.
\label{eq:truncated-distance}
\end{equation}
For simplicity, and without causing ambiguity, we will use \(D^\star(s,g)\) to denote \(D_{T_{\max}}(s,g)\) in the following. For numerical stability we also consider a normalized distance $\tilde D(s,g)=D_{T_{\max}}(s,g)/T_{\max}$ so that $\tilde D$ typically lies in $[0,1]$.

\textbf{Learning objectives.}
Our goal is to learn, solely from $\mathcal{D}$,
\begin{enumerate}
\item a \emph{Trajectory Length Predictor} $f_\theta:\mathcal{S}\times\mathcal{S}\to\mathbb{N}_{\ge 0}$ that approximates $D^\star$ (or its normalized variant $\tilde{D}$) using only state information; and
\item a \emph{variable-length diffusion planner} that, given $(s,g)$ and a target trajectory length $L\in\mathbb{N}$, generates a trajectory of length $L$ from $s$ toward $\mathcal{G}_\varepsilon(g)$.
\end{enumerate}
Both modules are trained on the same offline trajectories but with distinct supervision: the length predictor targets shortest-step reachability, while the planner learns to synthesize feasible sequences under randomly cropped lengths and later inferences with the predicted horizon. It is worth noting that, in our setting, since the initial state is included, the trajectory length equals the step distance plus one. Thus, the Length Predictor approximates the shortest step distance $D^\star$, while the actual trajectory length is $D^\star+1$.

\section{Preliminaries on Diffusion Models for Trajectory Planning} 
\label{sec:diffusion} 
Diffusion-based planners cast planning as conditional trajectory generation by learning the offline trajectory distribution $q(\bs{\tau}^0)$ induced by the environment dynamics and behavior in a dataset, and then sampling trajectories that satisfy task constraints (e.g., start/goal) at test time~\cite{janner2022planning,ajay2022conditional}. Here $\bs{\tau}^0 \in \mathbb{R}^{L\times d}$ denotes a noise-free trajectory of length $L$ (stacked states or state–action pairs). We adopt the convention that \emph{superscripts} index diffusion time and \emph{subscripts} index trajectory time; for instance, $\bs{\tau}^i_t$ is the $t$-th element of the noisy trajectory at diffusion step $i$.

\textbf{Forward (diffusion) process.}
A trajectory is gradually corrupted by Gaussian noise through a fixed Markov chain,
\begin{equation}
q(\bs{\tau}^{i}\mid \bs{\tau}^{i-1}) \;=\; \mathcal{N}\!\bigl(\bs{\tau}^{i};\, \sqrt{1-\beta_i}\,\bs{\tau}^{i-1},\, \beta_i \mathbf{I}\bigr), \; i=1,\dots,N,
\end{equation}
with variance schedule $\{\beta_i\}_{i=1}^N$. The marginal at step $i$ admits the closed form:
\begin{equation}
\bs{\tau}^{i} \;=\; \sqrt{\bar{\alpha}_i}\,\bs{\tau}^{0} \;+\; \sqrt{1-\bar{\alpha}_i}\,\bs{\varepsilon}, 
\; \bs{\varepsilon}\sim\mathcal{N}(\mathbf{0},\mathbf{I}), 
\; \bar{\alpha}_i=\prod_{j=1}^{i}(1-\beta_j).
\end{equation}

\textbf{Reverse (denoising) process.}
Sampling proceeds by reversing the diffusion with a parameterized Gaussian,
\begin{equation}
p_{\theta}(\bs{\tau}^{i-1}\mid \bs{\tau}^{i}) \;=\; \mathcal{N}\!\bigl(\bs{\tau}^{i-1};\, \bs{\mu}_{\theta}(\bs{\tau}^{i}, i),\, \bs{\Sigma}^{i}\bigr),
\end{equation}
where $\bs{\Sigma}^{i}$ is typically fixed and the mean is implemented via \emph{noise prediction}~\cite{ho2020denoising}:
\begin{equation}
\bs{\mu}_{\theta}(\bs{\tau}^{i}, i) \;=\; 
\frac{1}{\sqrt{\alpha_i}}\!\left(\bs{\tau}^{i} - \frac{1-\alpha_i}{\sqrt{1-\bar{\alpha}_i}}\, \bs{\varepsilon}_{\theta}(\bs{\tau}^{i}, i)\right),
\; \alpha_i = 1-\beta_i.
\end{equation}
The model $\bs{\varepsilon}_{\theta}$ is trained to predict the injected noise using the simplified objective
\begin{equation}
\mathcal{L}(\theta) \;=\; \mathbb{E}_{i,\bs{\varepsilon},\bs{\tau}^{0}}\!\left[\, \bigl\| \bs{\varepsilon} - \bs{\varepsilon}_{\theta}(\bs{\tau}^{i}, i) \bigr\|_2^2 \,\right].
\end{equation}

\textbf{Planning as conditioning.}
At test time, planning reduces to sampling a clean trajectory $\bs{\tau}^{0}$ consistent with task constraints by running the reverse chain from Gaussian noise. In practice, constraints such as start and goal can be enforced during the reverse process by simple conditioning operators (e.g., clamping the first/last state at each reverse step), yielding a trajectory that respects the specified boundary conditions while remaining on the learned data manifold.

\textbf{Generate Action Sequence.}  
Following mainstream practice, we use the diffusion model exclusively to generate state sequences. Since action sequences are typically high-frequency and less smooth~\cite{ajay2022conditional}, we recover them either offline via an inverse dynamics model~\cite{agrawal2016learning}, or online using a simple PD controller that tracks the generated state sequence, as in the implementation of~\cite{janner2022planning}.

\begin{figure*}[tbp]
  \centering
  \includegraphics[width=0.8\textwidth]{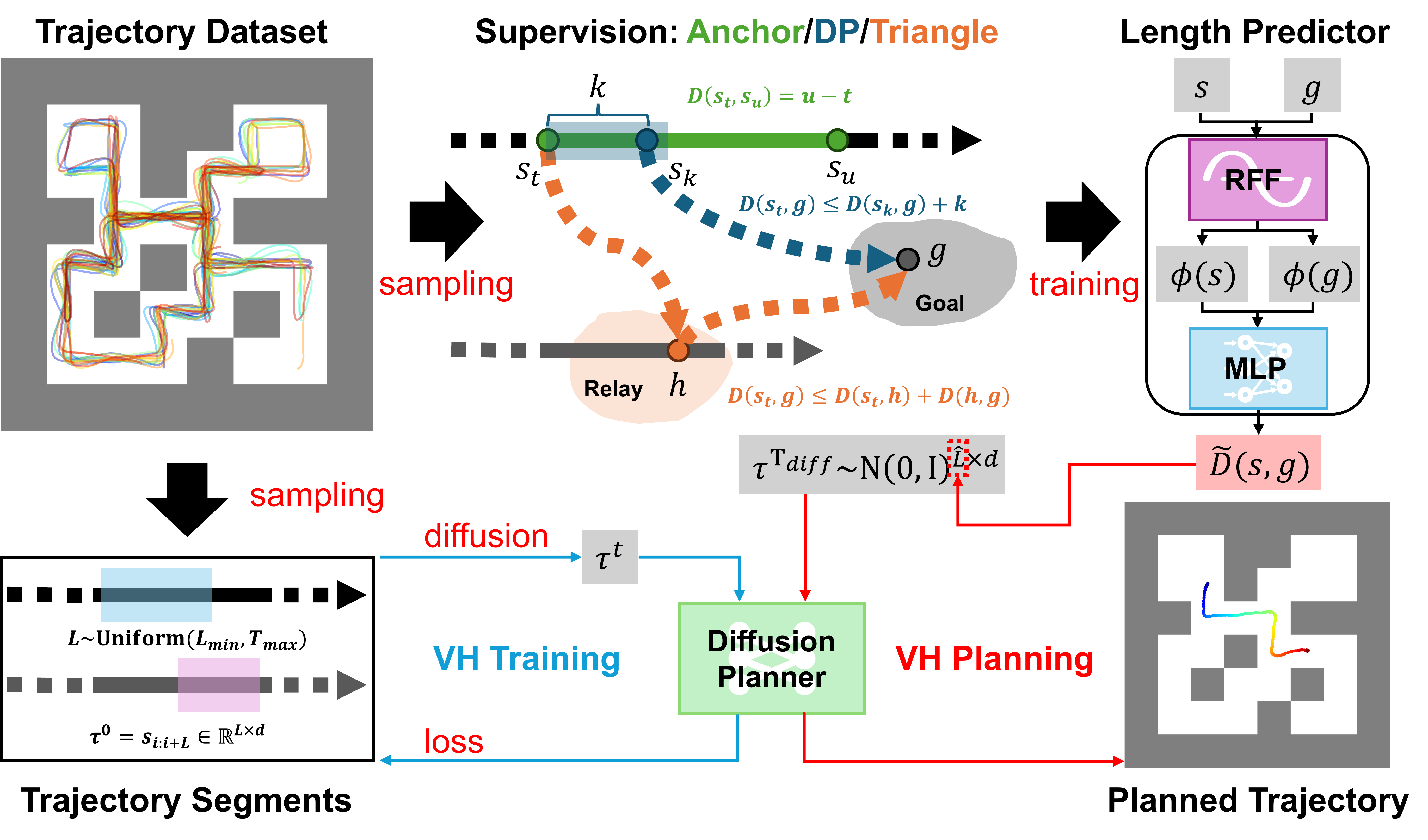}
  \caption{\textbf{\vhd Pipeline.} \lenpred estimates an instance-specific horizon, which sets the length of the initial noise for the \diffplan. Both modules are trained from the same offline trajectories; \lenpred uses dataset-derived hybrid supervision, and \diffplan uses random-length trajectory crops.}
  \label{fig:overview}
\end{figure*}

\section{Our Methods}
% =========================================================
\subsection{Overview of the \vhd Framework}
\label{sec:overview}
Our \method framework, illustrated in Fig.~\ref{fig:overview}, treats the planning horizon as a learned, instance-specific quantity while preserving the original diffusion planning backbone.
The framework consists of two key components: a \lenpred and a \diffplan.  Specifically, given a start–goal pair \((s,g)\), \lenpred estimates the shortest-step distance and produces a horizon \(\hat L\); the planner then runs the standard reverse diffusion with an initial noise trajectory of length \(\hat L\), enforcing start/goal via a simple conditioning operator. Training reuses the same offline dataset with different processing: \lenpred learns from hybrid supervisions constructed from trajectories (Sec.~\ref{sec:length-predictor}), and \diffplan learns from random-length trajectory crops (Sec.~\ref{sec:planner}). This design achieves both \emph{adaptivity} (mitigating length mismatch) and \emph{simplicity} (requiring no architectural changes to the planner).

\subsection{Trajectory Length Predictor}
\label{sec:length-predictor}
\textbf{Purpose.} Within \method, the \lenpred serves as the horizon selection mechanism of \method. Its role is to supply, for each start-goal pair $(s,g)$, an instance-specific planning horizon that reflects the underlying geometric and dynamical reachability. A reliable horizon is essential for the downstream \diffplan, which conditions its sampling length on this estimate and therefore benefits directly from accurate horizon calibration. Considering that even under fixed discrete-time dynamics $s_{t+1}=f(s_t,a_t)$, there typically exist many feasible trajectories between the same $(s,g)$ with markedly different lengths, offline datasets further exacerbate this variability since demonstrations often contain dithering, detours, and repeated corrections. As a result, the empirical trajectory lengths between similar state pairs are dispersed and typically biased upward relative to the true geometric difficulty. Therefore, instead of modeling the mean length or the full empirical length distribution, which is heavily skewed by behavior artifacts, we focus on the \emph{shortest-step} distance $D^\star(s,g)$ as a faithful surrogate for reachability and efficiency. This design aligns the predictor with the planner’s objective of generating efficient trajectories and decouples horizon estimation from the idiosyncrasies of the behavior policy.

\textbf{Data limitations and hybrid supervision.}
The main challenge in learning the \lenpred lies in the limited coverage of the offline trajectory dataset. Because the corpus consists of finitely many episodes and cannot enumerate all possible $(s,g)$ pairs, exact labels are abundant for state pairs that co-occur within the same trajectory, but scarce for \emph{cross-trajectory} pairs that never appear together. To address this, we adopt a hybrid supervision strategy that combines exact intra-trajectory supervision with conservative cross-trajectory constraints. Specifically, we employ three complementary supervisory signals, all defined in the normalized domain and clipped to $[0,1]$:

\begin{itemize}[leftmargin=*]
  \item Exact anchors (intra-trajectory labels). For episode $i$ of length $L_i$ and time $t$, any $u \in \{t{+}1,\dots,\min(t{+}T_{\max},\,L_i{-}1)\}$ yields an anchor pair $(s_t^i,s_u^i)$ with label $k=u{-}t$. If $s_t^i \in \mathcal{G}_\varepsilon(s_u^i)$ we also include the zero-length anchor $k=0$. Anchors supervise $\tilde D$ via the target
  $\min\{k/T_{\max},\,1\}$.
  \item DP upper bounds via $k$-step links (consistency). For a general pair $(s,g)$ without a direct hit, we sample $k \in \mathcal{K}$ (e.g., $\{1,2,4,8\}$)
  and define the $k$-step successor $s_k$ along the same episode as $s$ (if available).
  The dynamic-programming inequality
  $D^\star(s,g) \;\le\; k \;+\; D^\star(s_k,g)$
  induces the conservative normalized target
  $\min\{\,k/T_{\max} + \tilde D(s_k,g),\,1\}$, which encourages
  \emph{consistency} with feasible $k$-step relays and avoids optimistic bias.

  \item Relay-based triangle upper bounds (cross-trajectory). To further propagate constraints beyond a single episode, we introduce relay states $h$ drawn from
  (i) the on-trajectory relay $s_k$ and (ii) \emph{semi-hard} candidates sampled from the
  minibatch or a global pool. The triangle inequality $D^\star(s,g) \;\le\; D^\star(s,h) \;+\; D^\star(h,g)$
  yields the normalized upper bound
  $\min\{\,\tilde D(s,h) + \tilde D(h,g),\,1\}$, which \emph{stitches} geometric
  information across trajectories and strengthens supervision on long-range, cross-trajectory pairs.
\end{itemize}

All three signals are combined in the training objective: anchors contribute pointwise targets, while the DP and triangle terms act as upper-bound penalties that steer estimates toward the shortest-step geometry.

\textbf{Mini-batch construction.}
Each training batch consists of tuples $(s,g,k,s_k,\mathrm{hit})$ obtained by sampling from the offline corpus with a goal-mixture policy and a bounded lookahead. Concretely, we first sample an episode $i$ and index $t$ to obtain $s\!=\!s_t^i$. A goal $g$ is then drawn from a three-way mixture: (i) \emph{endpoint} states of each episode, (ii) \emph{local-future} states from the same episode within a window clipped by $T_{\max}$, and (iii) \emph{global} states drawn from a pool over the dataset. If $g$ lies in the $\ell_\infty$-tolerance neighborhood of some future state $s_u^i$ with $t{<}u{\le}\min(t{+}T_{\max},L_i{-}1)$, we mark $\mathrm{hit}{=}\,1$, set $k{=}u{-}t$, and take $s_k{=}s_u^i$ (including the zero-length case when $s_t^i\in\mathcal{G}_\varepsilon(s_u^i)$). Otherwise, we mark $\mathrm{hit}{=}\,0$, sample $k\!\in\!\mathcal{K}$ (e.g., $\{1,2,4,8\}$), and define $s_k$ as the $k$-step successor along episode $i$ when available (clipping to the episode end if necessary). Relay candidates $h$ used by the triangle penalty are drawn either as $s_k$ or via semi-hard mining~\cite{khosla2020supervised} from the current minibatch/global pool.

\textbf{Loss functions.}
Having established three complementary sources of supervision, we now unify them into a single learning objective. All quantities are expressed in the normalized domain $\tilde D=\min\{D^\star,T_{\max}\}/T_{\max}\in[0,1]$ so that targets and penalties live on a common scale, and only \emph{violations} of structurally valid upper bounds are penalized. This construction turns supervision into soft geometric constraints that bias the estimator toward the shortest-step geometry while preserving optimization stability.

Let $f_\theta:\mathcal{S}\times\mathcal{S}\!\to\![0,1]$ be the predictor and $\bar\theta$ an EMA copy used to form targets. We write $[x]_+=\max\{x,0\}$ and use a Huber penalty~\cite{huber1992robust} with threshold $\kappa>0$.

\emph{Primary Regression (Anchors and DP Targets).} For each batch element $(s,g,k,s_k,\mathrm{hit})$, anchors ($\mathrm{hit}=1$) provide an exact normalized target $\min\{1,k/T_{\max}\}$; otherwise we invoke the DP upper bound to form
\begin{equation}
\tilde y(s,g)=
\begin{cases}
\min\{1,\;k/T_{\max}\}, & \mathrm{hit}=1,\\[0.2em]
\min\{1,\;k/T_{\max}+f_{\bar\theta}(s_k,g)\}, & \mathrm{hit}=0.
\end{cases}
\label{eq:target}
\end{equation}
The pointwise term encourages $f_\theta$ to match these conservative targets:
\begin{equation}
\mathcal{L}_{\mathrm{TD}}
=\frac{1}{|\mathcal{B}|}\sum_{(s,g)\in\mathcal{B}}
\mathrm{Huber}\bigl(f_\theta(s,g)-\tilde y(s,g)\bigr),
\end{equation}
where $\mathrm{Huber}(\cdot)$ denotes the Huber loss\cite{huber1992robust}.

\emph{Consistency with $k$-Step Relays.} Regression alone does not guarantee respect of the DP structure. We therefore further penalize \emph{violations} of the DP upper bound, yielding a one-sided (hinge) consistency term:
\begin{equation}
\mathcal{L}_{\mathrm{cons}}
=\frac{1}{|\mathcal{B}|}\sum_{(s,g)\in\mathcal{B}}
\Bigl[
f_\theta(s,g)-\min\{1,\;k/T_{\max}+f_{\bar\theta}(s_k,g)\}
\Bigr]_+^{\,2}.
\label{eq:cons}
\end{equation}

\emph{Triangle Relays Across Trajectories.} To transport information beyond a single episode and to ``stitch'' long-range geometry, we add a relay-based triangle penalty using states $h$ drawn from $s_k$ and semi-hard candidates in the minibatch/global pool:
\begin{equation}
\mathcal{L}_{\triangle}
=\frac{1}{|\mathcal{B}|}\sum_{(s,g)\in\mathcal{B}}
\Bigl[
f_\theta(s,g)-\min\{1,\;f_{\bar\theta}(s,h)+f_{\bar\theta}(h,g)\}
\Bigr]_+^{\,2}.
\label{eq:triangle}
\end{equation}

\emph{Stabilizers.} Two soft constraints regularize trivial solutions and cap violations:
\begin{equation}
\mathcal{L}_{\mathrm{bdry}}=\frac{1}{|\mathcal{B}|}\sum_{g} f_\theta(g,g)^2,
\mathcal{L}_{\mathrm{clip}}=\frac{1}{|\mathcal{B}|}\sum_{(s,g)\in\mathcal{B}}
\bigl[f_\theta(s,g)-1\bigr]_+^{\,2}.
\label{eq:stabilizers}
\end{equation}

\emph{Composite objective.} The final loss is a weighted sum,
\begin{equation}
\begin{aligned}
\mathcal{L}
=\mathcal{L}_{\mathrm{TD}}
+\lambda_{\mathrm{cons}}\,\mathcal{L}_{\mathrm{cons}}
+\lambda_{\triangle}\,\mathcal{L}_{\triangle} \\
+\lambda_{\mathrm{bdry}}\,\mathcal{L}_{\mathrm{bdry}}
+\lambda_{\mathrm{clip}}\,\mathcal{L}_{\mathrm{clip}},
\label{eq:total-loss}
\end{aligned}
\end{equation}
with weights as tunable hyper parameters. 

\textbf{Models.}
We parameterize a scalar regressor \(f_\theta:\mathcal{S}\times\mathcal{S}\!\to\![0,1]\) that approximates the normalized shortest-step distance \(\tilde D(s,g)\) using \emph{states only}, so the estimate reflects geometric reachability rather than behavior-specific actions and remains drop-in compatible with existing diffusion planners. Each state is embedded with randomized Fourier features (RFF) \cite{tancik2020fourier} to provide a rich stationary basis; specifically, we encode \(x\) as \(\Phi(x)=[\sin(2\pi xB),\,\cos(2\pi xB),\,x]\) with a fixed Gaussian matrix \(B\), and form the joint representation \(z(s,g)=[\Phi(s),\,\Phi(g),\,\Phi(s)-\Phi(g)]\) where the relative channel \(\Phi(s)-\Phi(g)\) is optional but helps local geometry when \(s\approx g\). A lightweight MLP with normalization and ReLU maps \(z(s,g)\) to a scalar, followed by a softplus head to enforce nonnegativity, yielding \(f_\theta(s,g)=\mathrm{softplus}(\mathrm{MLP}(z(s,g)))\). The architecture is intentionally minimal and modality-agnostic, so stronger encoders (e.g., CNNs for pixels) can be substituted without altering the objective or the planner interface.

\textbf{Training details.}
To stabilize the training procedure, we employ EMA targets for \eqref{eq:target} and gradient clipping, and we adopt a curriculum that \emph{begins anchor-heavy} and \emph{gradually} increases the proportion and difficulty of cross-trajectory constraints since anchors provide relatively low-variance labels that quickly establish a reliable local geometry, whereas DP and triangle terms are higher-variance but expand coverage.

\emph{Phase I: Anchor Warm-Start.}
During the initial updates, sampling strongly favors intra-trajectory anchors, with a small link set $\mathcal{K}$ and rare relay use. The loss weights emphasize the primary regression, so optimization is dominated by exact or near-exact targets. This phase rapidly calibrates $\tilde D$ on relatively short ranges without introducing long-range inconsistencies.

\emph{Phase II: DP Expansion.}
Once the anchor error plateaus, we increase the share of unanchored pairs, enlarge $\mathcal{K}$, and activate the one-sided consistency term with a gentle ramp. Endpoint and global goals are sampled more frequently, which exposes the model to longer links while keeping the majority of targets anchored.

\emph{Phase III: Relay/Triangle Strengthening.}
In the later stage, we further enrich long-range coverage by raising the relay probability and enabling semi-hard mining to moderately tighten the triangle bound. Sampling shifts toward a balanced mix of local and global goals, so that cross-trajectory stitching becomes the primary driver of improvements while anchors continue to anchor calibration.

\subsection{Variable-Horizon Diffusion Planners}
\label{sec:planner}

\textbf{Key design.} 
In line with the principle of lightweight design, we adopt the diffusion planner from \cite{janner2022planning} as the backbone of our diffusion module, and we keep this backbone unchanged, explicitly \emph{without} introducing any horizon-conditioning input. Variable length is achieved by (i) training the same backbone on \emph{random-length sub-trajectories}, enabling it to internalize a distribution over lengths, and (ii) controlling the generated length at test time by \emph{modifying the shape of the initial noise trajectory (and sampler horizon)}. Following \cite{janner2022planning}, boundary conditioning is re-applied at every reverse step to enforce the endpoints consistently during sampling. This design renders the module drop-in compatible with existing diffusion planners, i.e., no architectural changes, no additional conditioning token, and no modification to the diffusion schedule.

\textbf{Training.}
As outlined in Sec.~\ref{sec:diffusion}, we adopt the standard $\epsilon$-prediction objective and cosine noise schedule, following \cite{janner2022planning}. The sole modification is \emph{variable-length exposure}: instead of fixing a training horizon, each update samples a random crop length (as highlighted in red in Algorithm~\ref{alg:train}). Concretely, given a demonstration $(s_{0:T})\!\sim\!\mathcal{D}$, we draw a crop length $L\!\sim\!\mathrm{Uniform}\{L_{\min},T_{\max}\}$ and a start index $i\!\in\!\{0,\dots,T{-}L\}$ to form a contiguous sub-trajectory $\tau^{0}\!=\!s_{i:i+L}\!\in\!\mathbb{R}^{L\times d}$ containing $L{-}1$ consecutive \emph{trajectory} steps. We then sample a \emph{diffusion} timestep $t\!\in\!\{1,\dots,T_{\mathrm{diff}}\}$, corrupt $\tau^{0}$ with the prescribed forward noising process, and train the denoiser $\mathrm{DM}_\phi$ to predict the injected noise with the usual mean-squared error. Apart from drawing $(i,L)$ at each update, all components (e.g., timestep sampling, noising, network architecture, and optimization) are identical to \cite{janner2022planning}. In practice, random-length crops prevent overfitting to a single horizon and substantially improve the planner’s ability to generate trajectories of different lengths at inference time.

\begin{algorithm}[tbp]
\caption{Variable-horizon diffusion training}
\label{alg:train}
\begin{algorithmic}[1]
\Require dataset $\mathcal{D}$, minimum/maximum crop lengths $L_{\min},T_{\max}$, diffusion schedule $\{\beta_t\}_{t=1}^{T_{\mathrm{diff}}}$
\Repeat
  \State sample a full demonstration $(s_{0:T})\sim\mathcal{D}$
  \State \textcolor{red}{sample crop length $L\sim\mathrm{Uniform}\{L_{\min},\,T_{\max}\}$ and start index $i\sim\{0,\dots,T{-}L\}$}
  \State \textcolor{red}{form sub-trajectory $\tau^0 \gets s_{i:i+L}\in\mathbb{R}^{L\times d}$}
  \State sample diffusion step $t\sim\{1,\dots,T_{\mathrm{diff}}\}$ and noise $\epsilon\sim\mathcal{N}(0,I)^{L\times d}$
  \State $\tau^t \gets \sqrt{\bar\alpha_t}\,\tau^0 + \sqrt{1{-}\bar\alpha_t}\,\epsilon$ \Comment{$\bar\alpha_t=\prod_{j=1}^{t}(1{-}\beta_j)$}
  \State $\hat\epsilon \gets \mathrm{DM}_\phi(\tau^t,\,t)$ \Comment{standard $\epsilon$-prediction}
  \State update $\phi$ with loss $\mathcal{L}=\lVert \epsilon-\hat\epsilon\rVert^2$
\Until{converged}
\end{algorithmic}
\end{algorithm}

\textbf{Inference (Planning).}
At test time the planning pipeline mirrors Diffuser \cite{janner2022planning} in all respects except for how we set the trajectory length (as highlighted in red in Algorithm~\ref{alg:infer}). Given a start–goal pair $(s,g)$, we first query the length predictor to obtain the step distance $\hat L$. In practice, $\hat L=\mathrm{clip}(\gamma\cdot (f_\theta(s,g)\cdot T_{\max}+1),L_{\min},T_{\max})$, where $\gamma$ is a user-specified scaling factor that provides a certain margin. We then sample an initial latent noisy plan of \emph{trajectory length} $\hat L$, $\tau^{T_{\mathrm{diff}}}\!\sim\!\mathcal{N}(0,I)^{\hat L\times d}$, and run the standard denoising steps. At each reverse step we compute the noise estimate $\hat\epsilon=\mathrm{DM}_\phi(\tau^{t},t)$ and apply the usual posterior update to obtain $\tau^{t-1}$. Immediately after each update, we enforce endpoint consistency by replacing the first and last states of trajectory with $s$ and $g$ respectively as in \cite{janner2022planning}. The overall process is shown in Algorithm~\ref{alg:infer}.

\begin{algorithm}[tbp]
\caption{Planning with \method}
\label{alg:infer}
\begin{algorithmic}[1]
\Require start $s$, goal $g$, length predictor $f_\theta$, diffusion planner $\mathrm{DM}_\phi$
\State \textcolor{red}{Predict trajectory length $\hat{L}$ with $f_\theta$}
\State \textcolor{red}{Sample initial plan $\tau^{T_{\mathrm{diff}}} \sim \mathcal{N}(0,I)^{\hat L\times d}$}
\For{$t{=}\,T_{\mathrm{diff}},\dots,1$}
  \State $\hat\epsilon \gets \mathrm{DM}_\phi(\tau^t,t)$;
  \;$\tau^{t-1}\gets \mathrm{posterior\_step}(\tau^t,\hat\epsilon,t)$
  \State $\tau^{t-1}_0 \gets s, \quad \tau^{t-1}_{\hat{L}-1} \gets g$
\EndFor
\State \Return trajectory $\tau^0$ of length $\hat L$
\end{algorithmic}
\end{algorithm}

\section{Experiments}
\subsection{Experimental Setup}
\label{sec:exp-setup}

\textbf{Environments.}
We evaluate our approach on five benchmarks covering both navigation and robot arm control: three D4RL ~\cite{fu2020d4rl} Maze2d tasks (\textit{umaze}, \textit{medium}, \textit{large}) with a point agent navigating a 2D maze; one OGBench~\cite{ogbench_park2025} AntMaze task (\textit{medium}) with an 8-DoF ant agent performing long-horizon navigation; and one OGBench \textit{cube} robot arm control task with a 6-DoF UR5e robot arm. In the \textit{cube} environment, we disregard the original cube reconfiguration task and focus solely on end-effector positioning. All experiments are based exclusively on the trajectory datasets provided by the respective benchmarks.

\textbf{Training protocol.}
Unless otherwise specified, both the length predictor and the diffusion planner are trained on the offline datasets of each environment. For the planner, our variable-horizon (VH) training follows the same diffusion objective as a standard diffusion planner, but replaces fixed-length trajectory segments with random-length sub-trajectories, as described in Sec.~\ref{sec:planner}. This contrasts with baseline planners trained exclusively on fixed-length segments.

\textbf{Baselines and variants.}
We compare our method (\method) against:  
(i) three \emph{fixed-horizon} diffusion planners (\textbf{FH-\(H\)}) trained and evaluated with fixed horizons $H\!\in\!\{H_1,H_2,H_3\}$, where $H_1,H_2,H_3$ are chosen per environment as shown in Table~\ref{tab:parameters}; and  
(ii) a \emph{fixed-train / variable-test} variant (\textbf{FH-LP}), trained on fixed-length trajectories but using the learned Length Predictor at inference time to set the horizon. This variant isolates the effect of VH training from that of horizon selection at test time.

\textbf{Execution protocols (controllers).}
As mentioned in Sec.~\ref{sec:diffusion}, following common practice, we use the diffusion planner solely to generate state sequences. The corresponding action sequences are then obtained by tracking the trajectories with either a controller or an inverse dynamics model. We adopt two standard control modes for execution:
\begin{enumerate}
\item \textbf{Single-shot tracking.} The planner is invoked once to produce a state trajectory; execution then applies either a PD controller or an inverse-dynamics model to track the trajectory. This mode targets settings where real-time responsiveness and control-step budget are critical.
\item \textbf{Replan-on-deviation.} Starting from the same initial plan, if the instantaneous tracking error exceeds a preset threshold during execution, the planner is re-invoked from the current state to produce a new trajectory toward the original goal. This mode targets accuracy-sensitive settings. To avoid excessively frequent replanning, the tracking error is only checked at a user-specified frequency.

\end{enumerate}
In both modes we adopt a strict synchronization protocol to better isolate planning quality: the \emph{planning timestep equals the control timestep}, and at each trajectory index the controller is allowed to generate \emph{exactly one} action before advancing to the next index. In Maze2D environments, we adopt a PD controller, whereas in the AntMaze and Cube environments, we utilize an inverse dynamics model to generate actions.

\begin{table}[tbp]
\caption{Key evaluation hyperparameters per environment. $H_1,H_2,H_3$ are fixed planning horizons (steps); $\varepsilon$ is the (normalized) goal tolerance measured in $\ell_\infty$.}
\label{tab:parameters}
\setlength{\tabcolsep}{6pt}
\centering
\footnotesize
\begin{tabular}{lcccc}
\toprule
Environment & $H_1$ & $H_2$ & $H_3$ & $\varepsilon$ \\
\midrule
Maze2d-umaze  & 64   & 128   & 192   & 0.04 \\
Maze2d-medium & 192   & 288   & 384   & 0.03 \\
Maze2d-large  & 256 & 384 & 512 & 0.03 \\
AntMaze       & 256 & 384 & 512  & 0.05 \\
Cube   & 32   & 64   & 128   & 0.03 \\
\bottomrule
\end{tabular}
\end{table}

\textbf{Test cases and evaluation metrics.}
For each environment we randomly generate $N_{\text{test}}=1000$ evaluation instances. Each instance specifies a random start state and a random goal state, sampled independently of the training data. We use two complementary metrics.

\emph{Success rate.}
An execution is deemed successful if the agent’s final state lies within a fixed, environment-specific error threshold $\varepsilon_{\text{env}}$ of the goal (shown in Table~\ref{tab:parameters}). We report the fraction of successful instances,
$
\mathrm{SR}
=\frac{1}{N_{\text{test}}}\sum_{i=1}^{N_{\text{test}}}
\mathbf{1}\!\left[\operatorname{dist}\!\bigl(s^{(i)}_{\text{final}},\,g^{(i)}\bigr)\le \varepsilon_{\text{env}}\right],
$
where $\operatorname{dist}(\cdot,\cdot)$ is the task-specific state-space distance and we choose the $\ell_\infty$ norm in our experiments. The same $\varepsilon_{\text{env}}$ is used across all methods within an environment.

\emph{Average executed steps.}
To quantify planning efficiency under our strict synchronization protocol (planning timestep equals control timestep, with exactly one control action per index), we measure the average number of control timesteps actually issued:
\(
\mathrm{AES}
=\frac{1}{N_{\text{test}}}\sum_{i=1}^{N_{\text{test}}}\ell^{(i)}.
\)
Here, $\ell^{(i)}$ denotes the realized trajectory length in control timesteps \emph{until termination}. Since execution terminates immediately upon entering the goal region, $\ell^{(i)}$ can be \emph{shorter} than the planned horizon (e.g., in the case of an early hit). Under single-shot tracking, this corresponds to
$\ell^{(i)}=\min\{\hat L^{(i)},\,t^{(i)}_{\text{hit}}\},$
where $t^{(i)}_{\text{hit}}$ is the first index $k$ such that $\|s_k^{(i)}-g^{(i)}\|_\infty\le\varepsilon_{\text{env}}$. Under replan-on-deviation, $\ell^{(i)}$ is defined as the sum of the lengths of all executed plan segments up to the first successful hit (or until failure if the maximum timestep is reached). This definition ensures that $\mathrm{AES}$ reflects the actual executed control effort, rather than the nominal plan length.

\subsection{Results and Analysis}
\label{sec:results}

\begin{table}[tbp]
\caption{Results on five environments. Top sub-row shows SR (\%)$\uparrow$, bottom sub-row shows AES (steps)$\downarrow$.
Env: U/M/L = Maze2d-\{umaze, medium, large\}, Ant = AntMaze. A $\dagger$ following a result denotes the best value, while a $\ddagger$ denotes the second-best value.
}
\label{tab:main-results}
\setlength{\tabcolsep}{2pt}
\renewcommand{\arraystretch}{1.05}
\footnotesize
\centering
\begin{tabularx}{\columnwidth}{l *{5}{Y}}
\toprule
\multirow{2}{*}{Method} & \multicolumn{5}{c}{Environments} \\
& U & M & L & Ant & Cube \\
\midrule
\mrow{VHD (SS)}       
  & 97.1\% $\dagger$ & 93.1\% $\ddagger$ & 91.0\% $\ddagger$ & 82.2\% $\dagger$ & 86.9\% $\ddagger$ \\
\cmidrule(lr){2-6}
  & 112.73  & 230.02 $\ddagger$ & 245.84 $\ddagger$ & 152.79 $\ddagger$  & 50.87 $\ddagger$ \\
\midrule
\mrow{FH+LP (SS)}    
  & 93.2\% & 88.8\% & 80.4\% & 72.1\% & 86.2\% \\
\cmidrule(lr){2-6}
  & 114.24  & 232.45  & 250.52  & 143.94 $\dagger$ & 52.98  \\
\midrule
\mrow{FH-$H_1$ (SS)} 
  & 63.3\% & 84.7\% & 79.8\% & 75.8\% & 74.2\% \\
\cmidrule(lr){2-6}
  & 55.47 $\dagger$  & 169.60 $\dagger$  & 221.17 $\dagger$ & 209.87  & 30.19 $\dagger$ \\
\midrule
\mrow{FH-$H_2$ (SS)} 
  & 95.5\% & 94.5\% $\dagger$ & 93.8\% $\dagger$ & 79.9\% $\ddagger$ & 84.9\% \\
\cmidrule(lr){2-6}
  & 100.75 $\ddagger$ & 243.81  & 319.31  & 307.32  & 56.73  \\
\midrule
\mrow{FH-$H_3$ (SS)} 
  & 96.0\% $\ddagger$ & 92.4\% & 85.9\% & 76.8\% & 87.0\% $\dagger$ \\
\cmidrule(lr){2-6}
  & 140.61  & 326.68  & 422.93  & 413.50  & 111.15  \\
\specialrule{0.9pt}{3pt}{3pt}
\mrow{VHD (RP)}       
  & 100\% $\dagger$ & 98.7\% $\dagger$ & 99.6\% $\dagger$ & 95.7\% $\dagger$ & 98.2\% $\dagger$ \\
\cmidrule(lr){2-6}
  & 139.96 $\dagger$ & 319.96 $\ddagger$ & 285.66 $\dagger$  & 518.31 $\dagger$ & 77.51 $\ddagger$ \\
\midrule
\mrow{FH+LP (RP)}    
  & 99.9\% $\ddagger$ & 95.0\% & 98.8\% $\ddagger$ & 94
  .0\%  & 97.4\% \\
\cmidrule(lr){2-6}
  & 159.92 $\ddagger$  & 390.86  & 358.94  & 664.86 $\ddagger$ & 90.59  \\
\midrule
\mrow{FH-$H_1$ (RP)} 
  & 88.6\% & 94.6\% & 98.9\% & 94.4\% $\ddagger$ & 97.8\% \\
\cmidrule(lr){2-6}
  & 238.82  & 305.98 $\dagger$  & 300.40 $\ddagger$ & 681.07  & 63.08 $\dagger$ \\
\midrule
\mrow{FH-$H_2$ (RP)} 
  & 99.4\% & 97.2\% $\ddagger$ & 99.6\% $\dagger$ & 91.6\% & 97.9\% $\ddagger$\\
\cmidrule(lr){2-6}
  & 163.23  & 372.10  & 427.02  & 994.62  & 87.47  \\
\midrule
\mrow{FH-$H_3$ (RP)} 
  & 99.0\% & 95.2\% & 96.2\% & 89.5\% & 97.9\% $\ddagger$\\
\cmidrule(lr){2-6}
  & 219.36  & 508.97  & 639.69  & 1227.04  & 139.20  \\
\bottomrule
\end{tabularx}
\end{table}

Table~\ref{tab:main-results} summarizes success rate (SR, higher is better) and average executed steps (AES, lower is better) across five benchmarks and two execution protocols: single-shot tracking (SS) and replan-on-deviation (RP).

% \textbf{Overall trends.}
% On long-horizon, geometrically diverse AntMaze, \method (\textbf{VH}) achieves the best SR under both SS and RP while keeping AES competitive, indicating that instance-adaptive horizons mitigate length mismatch effectively. On Maze2d, which concentrates around a narrower range of instance difficulties, a \emph{mid-range fixed} horizon (\textbf{FH-$H_2$}) often attains the highest SR in the SS setting, whereas \textbf{VH} attains substantially lower AES with only a modest SR drop. Under the RP setting, \textbf{VH} achieves the highest SR in nearly all environments while also attaining the shortest AES, significantly outperforming other methods. 

\textbf{Overall trends.}
Across the five benchmarks, \method is consistently competitive. Under \textbf{RP}, \textbf{VHD(RP)} attains the best success rate in all environments while also achieving the lowest or second–lowest AES, indicating that horizon adaptivity dominates when plans are periodically revised. Under \textbf{SS}, Maze2d often favors a \emph{mid–range} fixed horizon (\textbf{FH-$H_2$}) for the very highest SR, yet \textbf{VHD(SS)} stays within a small margin (usually 1–3\%) while yielding shorter executed trajectories than the SR–winning fixed baseline. Notably, \textbf{VHD(SS)} achieves the top SR on \emph{Umaze} and \emph{AntMaze} with competitive AES, highlighting the advantage of instance–adaptive horizons when difficulty varies widely.

\textbf{Single-shot tracking (SS).}
With a single planning call and exactly one control action per index, \textbf{VHD(SS)} trades at most a modest SR gap to the best fixed horizon planner on Maze2d or Cube for clear efficiency gains. This pattern arises because many Maze2d or Cube test pairs may cluster around a common geometric scale that happens to align with certain fixed horizon, giving it an unusual advantage; by contrast, \textbf{VHD(SS)} must generalize across lengths and can under–predict on a subset of such cases. The strict one–step–per–index control mode also makes shorter plans harder to track (larger waypoint gaps and fewer control steps), slightly penalizing SR. Even so, \textbf{VHD(SS)} achieves the best SR on \emph{Umaze} and \emph{AntMaze}, and the second-best SR on the other environments, indicating that horizon adaptivity is beneficial across all scenarios. Moreover, it alleviates the burden of manually selecting horizons for each environment and each case.

% \textbf{Single-shot tracking (SS).}
% With a single call to the planner and strictly one control action per index, \textbf{VH} produces shorter plans (lower AES) but can yield slightly lower SR than \textbf{FH-$H_2$} on Maze2d. Two factors help explain this gap. First, random start-goal pairs in Maze2d often cluster around a characteristic geometric scale, so a single mid-range horizon aligns with a large fraction of test cases; by contrast, \textbf{VH} must generalize across lengths and can under-predict in a subset of instances. Second, shorter plans impose larger inter-waypoint displacements under the strict ``one-step-per-index'' controller, which can be harder to track, thus reduce SS success. This behavior is consistent with the efficiency-robustness trade-off.

\begin{figure}[tbp]
  \centering
  \includegraphics[width=0.48\textwidth]{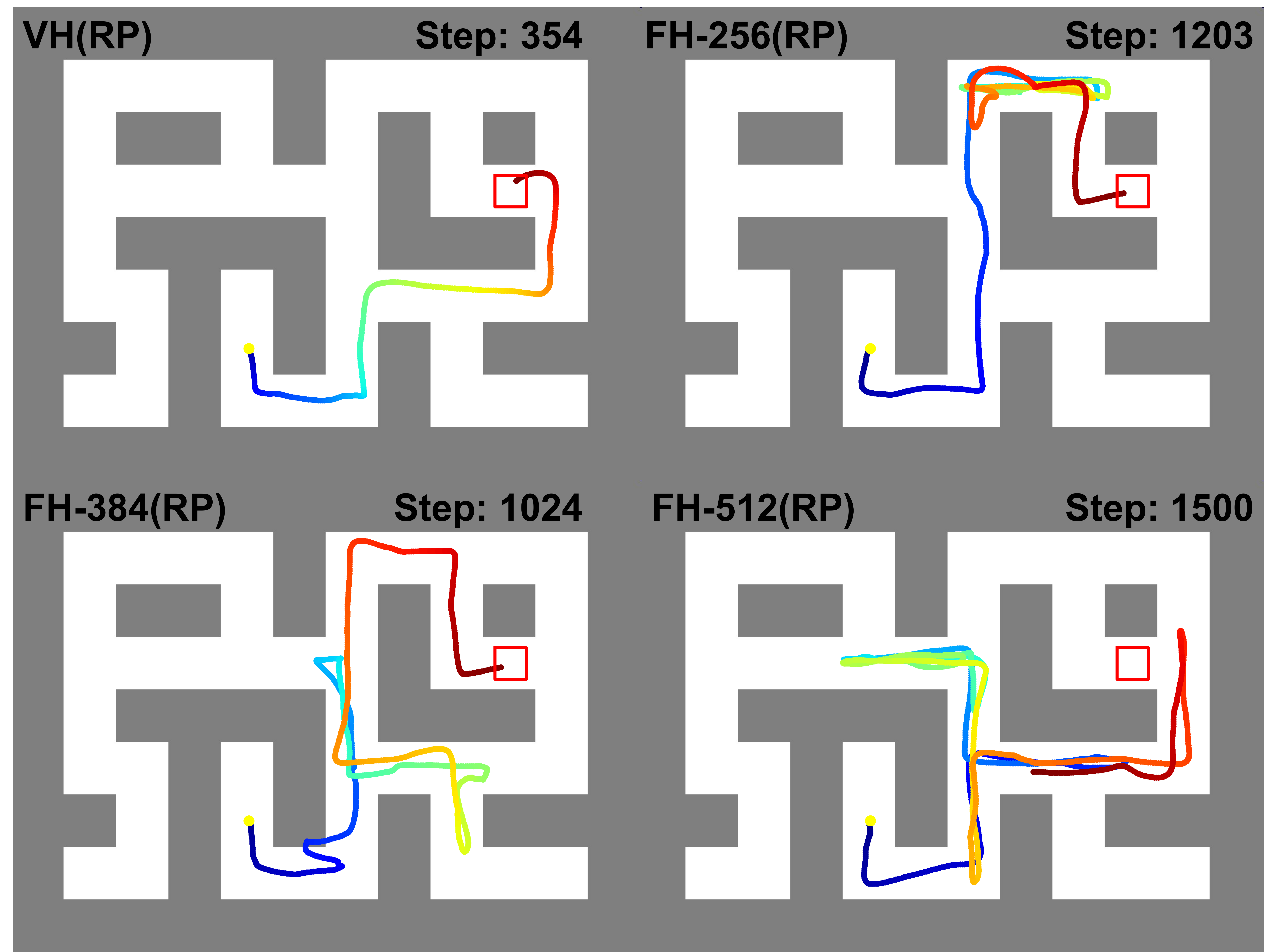}
  \caption{\textbf{Horizon mismatch under replanning (Maze2d-Large).}
  Executed trajectories for one start–goal pair with four planners:
  \textbf{VHD(RP)} (top-left), \textbf{FH-256(RP)} (top-right), \textbf{FH-384(RP)} (bottom-left), and \textbf{FH-512(RP)} (bottom-right).
  Start (yellow) and goal region (red) are shown.
  \textbf{VHD(RP)} adapts segment length to residual distance and yields a direct route.
  \textbf{FH-256(RP)} detours near the goal where shorter segments are required.
  \textbf{FH-384/512(RP)} detour early due to overly long segments; \textbf{FH-512(RP)} exceeds the step budget (1500) and fails to enter the goal region.}
  \label{fig:case1}
\end{figure}

\textbf{Replan-on-deviation (RP).}
Replanning is intrinsically \emph{horizon-sensitive}: far from the goal, longer corrective segments are desirable, whereas near the goal, shorter segments avoid overshoot and dithering. \textbf{VHD(RP)} adapts the horizon at each replan to the current residual distance, aligning horizon with task phase. Fixed-horizon planners cannot adjust, yielding unnecessary successive replans when the horizon is too short and inflated executed steps (and detours) when it is too long. Fig.~\ref{fig:case1} visualizes this effect on a \textit{Maze2d-Large} instance: \textbf{VHD(RP)} (top-left) produces a direct path with few replans; \textbf{FH-256(RP)} (top-right) is reasonable early but detours near the goal where shorter segments are needed; \textbf{FH-384(RP)} (bottom-left) and \textbf{FH-512(RP)} (bottom-right) exhibit early-stage detours due to overly long segments, with \textbf{FH-512(RP)} failing to reach the goal within the step limit. These behaviors explain the higher SR and shorter AES of \method under RP in Table~\ref{tab:main-results}.

\textbf{Effect of variable-length training.} The fixed-train/variable-test variant (\textbf{FT+LP}) which uses a Length Predictor at inference but is trained on fixed-length crops consistently underperforms \textbf{VHD}. This gap isolates the benefit of variable-length exposure during training: by learning from random-length sub-trajectories, the planner acquires invariances to trajectory length that are essential for reliable generation once the horizon is set by the predictor.

\textbf{Discussion.}
\method delivers a superior efficiency–reliability trade-off by eliminating length mismatch through instance-adaptive horizons. Across tasks with broad variability (e.g., AntMaze), it attains the highest success while executing markedly shorter trajectories; under RP, this advantage is amplified as horizons adapt at each replan. In settings where a single mid-range horizon happens to align with many cases (e.g., Maze2d SS), fixed-$H$ can match or slightly exceed success but only by incurring longer paths. Crucially, variable-length training is key as it enables a single backbone to generalize across horizons. A primary limitation is coverage: purely offline datasets may undersample rare start–goal pairs or long-range relays, which can weaken supervision and bias horizon estimates; a related issue is out-of-distribution generalization when test pairs depart from the dataset’s geometric support. These are well-known challenges in offline settings. Our hybrid supervision (anchors, DP upper bounds, and relay-based triangle constraints) partially mitigates them by propagating constraints across trajectories, but cannot fully close the gap without additional signal. In practice this implies that the Length Predictor’s estimates may not always match the instance-optimal horizon; nevertheless, the diffusion planner trained with variable-length crops exhibits strong length generalization, so the combined system still produces appropriately scaled, high-success plans in most evaluated scenarios. Future work will explore uncertainty-aware length prediction, coverage-aware sampling, and stitching-based data augmentation~\cite{li2024diffstitch} to improve robustness under limited coverage and OOD shifts.

\textbf{Practical guidance.}
From a deployment perspective, the choice of control protocol should reflect task constraints. When the application is sensitive to the \emph{number of executed steps} or to \emph{real-time latency} (e.g., strict budgets that preclude replanning), \textbf{Single-Shot (SS)} execution is preferable. In this regime, \method typically attains near optimal success rates while delivering \emph{markedly shorter} executed trajectories, without training multiple models, tuning horizon grids, or collecting extra data. In scenarios where accuracy is paramount and occasional replans are acceptable, we recommend \textbf{Replan-on-Deviation (RP)}. Because RP is highly horizon-sensitive, variable-length planning confers a stronger advantage: \method adapts segment length to the residual distance at each replan, which \emph{simultaneously} boosts success and reduces executed steps relative to fixed horizon planners. Overall, across both SS and RP, \method offers an economical, lightweight default—one backbone and one training recipe that generalize across horizons \emph{without} altering the network architecture—making it easy to drop into existing diffusion planning pipelines with minimal code changes. Moreover, the Length Predictor yields an interpretable horizon estimate that can act as a heuristic for higher-level planners (e.g., waypoint selection, graph search, or task-and-motion planning), guiding replan timing and control-step allocation~\cite{liu2025zero}.

% \textbf{Planning performance.} \method improves success and SPL over FHDP-Single across tasks and is competitive with, or surpasses, FHDP-Grid under matched total budgets. Gains are largest for instances whose optimal horizons deviate from the dataset mode.

% \textbf{Robustness to horizon mismatch.} Sensitivity curves (success/SPL vs.\ scaling $\lambda\hat L$) show substantially flatter profiles for \method than fixed-horizon baselines, confirming reduced brittleness to length errors.

% \textbf{Length generalization.} On \emph{unseen} length ranges, random-length training plus noise-shape control yields consistent performance, while fixed-horizon planners degrade markedly.

% =========================================================
\section{Conclusion}
We presented \method, a modular framework that elevates horizon selection from a fixed hyperparameter to a learned, instance-specific variable via a Length Predictor, and realizes variable horizon planning in a standard diffusion planner by setting the initial noise length with no architectural changes. Across navigation and robot arm control benchmarks, \method improves the success–efficiency trade-off, particularly under replan-on-deviation control protocol and in tasks with broad length variability. Future work includes uncertainty-aware length prediction and risk-sensitive planning, joint or end-to-end training of the predictor and planner, active data augmentation to improve coverage, tighter coupling with adaptive low-level controllers, and deploying the framework on real robots and more diverse long-horizon tasks.

% \section*{APPENDIX}

% Appendixes should appear before the acknowledgment.

% \section*{ACKNOWLEDGMENT}

% The preferred spelling of the word “acknowledgment” in America is without an “e” after the “g”. Avoid the stilted expression, “One of us (R. B. G.) thanks . . .”  Instead, try “R. B. G. thanks”. Put sponsor acknowledgments in the unnumbered footnote on the first page.

%%%%%%%%%%%%%%%%%%%%%%%%%%%%%%%%%%%%%%%%%%%%%%%%%%%%%%%%%%%%%%%%%%%%%%%%%%%%%%%%

\bibliographystyle{IEEEtran}
\bibliography{reference}{}

\end{document}